\documentclass{article}
\usepackage{ijcai21}

\usepackage{times}
\usepackage{soul}
\usepackage{url}
\usepackage[hidelinks]{hyperref}
\usepackage[utf8]{inputenc}

\usepackage{graphicx}
\usepackage{amsmath}
\usepackage{amsthm}
\usepackage{booktabs}
\usepackage{subcaption}

\usepackage[linesnumbered]{algorithm2e}
\usepackage[font=small,labelfont=bf]{caption}

\title{Bayesian Meta-reinforcement Learning for Traffic Signal Control}

\author{%
  Yayi Zou, Zhiwei Qin
    \\
   \textsc{Didichuxing AI Lab}
}

\everymath{\textstyle}

\begin{document}

\maketitle

\begin{abstract}
In recent years, there has been increasing amount of interest around meta reinforcement learning methods for traffic signal control, which 
have achieved better performance compared with traditional control methods. However, previous methods lack robustness in adaptation and stability in training process in complex situations, which largely limits its application in real-world traffic signal control. In this paper, we propose a novel value-based Bayesian meta-reinforcement learning framework BM-DQN to robustly speed up the learning process in new scenarios by utilizing well-trained prior knowledge learned from existing scenarios. This framework is based on our proposed fast-adaptation variation to Gradient-EM Bayesian Meta-learning and the fast-update advantage of DQN, which allows for fast adaptation to new scenarios with continual learning ability and robustness to uncertainty. The experiments on restricted 2D navigation and traffic signal control show that our proposed framework adapts more quickly and robustly in new  scenarios than previous methods, and specifically, much better continual learning ability in heterogeneous scenarios. 
\end{abstract}

\section{Introduction}

Traffic signal control in intersections takes important role in our everyday life. Efforts have been made to design systems that can react to the feedback from the environment in order to save travel time of vehicles passing through the intersections. However, traditional traffic control systems, e.g., SCATS \cite{pr1992scats}, use handcraft traffic signal plans which make them hard to find optimal control solutions in dynamic situations of intersections. Since intersection traffic signal control can be well modeled as a Markov Decision Process (MDP), it is naturally to apply reinforcement techniques to this problem, especially given the growing available traffic data (e.g., surveillance camera data) and development in deep reinforcement learning (DRL) \cite{lillicrap2015continuous}. In recent years, DRL-based methods in traffic signal control have shown better performance than traditional methods \cite{wei2018intellilight,wei2019survey,van2016coordinated}. 

However, the training process of DRL involves a lot of exploration/exploitation and thus needs a large amount of data and learning time to achieve superior performance, which is not feasible in many practical situations. Therefore, recent studies apply Meta-learning into traffic signal control \cite{zang2020metalight} which utilized the common knowledge of existing tasks(referres as meta-knowledge) and learn to quickly adapt to new tasks. 

Nevertheless, our empirical studies show that previous methods lack robustness in adaptation and stability in training process under more complicate settings where the available meta-knowledge from existing tasks may not be sufficient for immediate quick adaptation. These drawbacks are crucial to transportation systems where bad explorations may result in severe traffic problems. This motivates us to develop a traffic control algorithm that are able to learn robustly with only a few samples. 

Our contributions come in two folds. First, we propose a Bayesian meta-learning algorithm BM-DQN with robust continual learning ability. Unlike previous methods, e.g. MetaLight \cite{zang2020metalight}, which learns an initial point as meta-knowledge, BM-DQN learns a prior distribution as meta-knowledge of previously learned tasks. When comes to new tasks, BM-DQN infers a task posterior based on the learned prior and data from that task. This Bayesian probabilistic foundation effectively mitigates the instability in training process and enhances its robustness in adaptation to a new task. BM-DQN alternatively performs two updates, individual update which adapts to each individual task and global update which aggregate the common knowledge learned from each individual adaptation. Specifically, we design a fast adaptation variation to effectively speeds up the adaptation process such that this meta-learning algorithm could work on DQN (since the higher update frequency in DQN requires faster adaptation ability than policy-based RL). 
Second, we apply BM-DQN to traffic signal control. Compared to previous methods, the Bayesian modeling adds robustness continual learning ability in the adaptation to new scenarios which is specifically important in traffic signal control. In real-world traffic signal control where only data of limited intersections is available, it is important to have robust continual learning ability especially when meta-knowledge is not sufficient yet. The success of robust continual learning on individual tasks will then help learn a better meta-knowledge. Thus, this design forms a positive feedback loop.

In experiment, we first test our proposed BM-DQN framework on a designed restricted 2D navigation problem to demonstrate its superiority in value-based RL. Then we conduct extensive experiments to evaluate its performance in traffic signal control on four real-world datasets proposed by \cite{zang2020metalight}. The results show that our proposed BM-DQN outperforms state-of-the-art baselines in traffic signal control. Specifically, in challenging settings, it demonstrates the superiority of Bayesian method in robust continual individual adaptation and stability during training process. 

\section{Markov Decision Process for Traffic Signals Control}

\begin{figure}[t]
\centering     
    \begin{subfigure}[b]{0.33\textwidth}
         \centering
         \includegraphics[width=\textwidth]{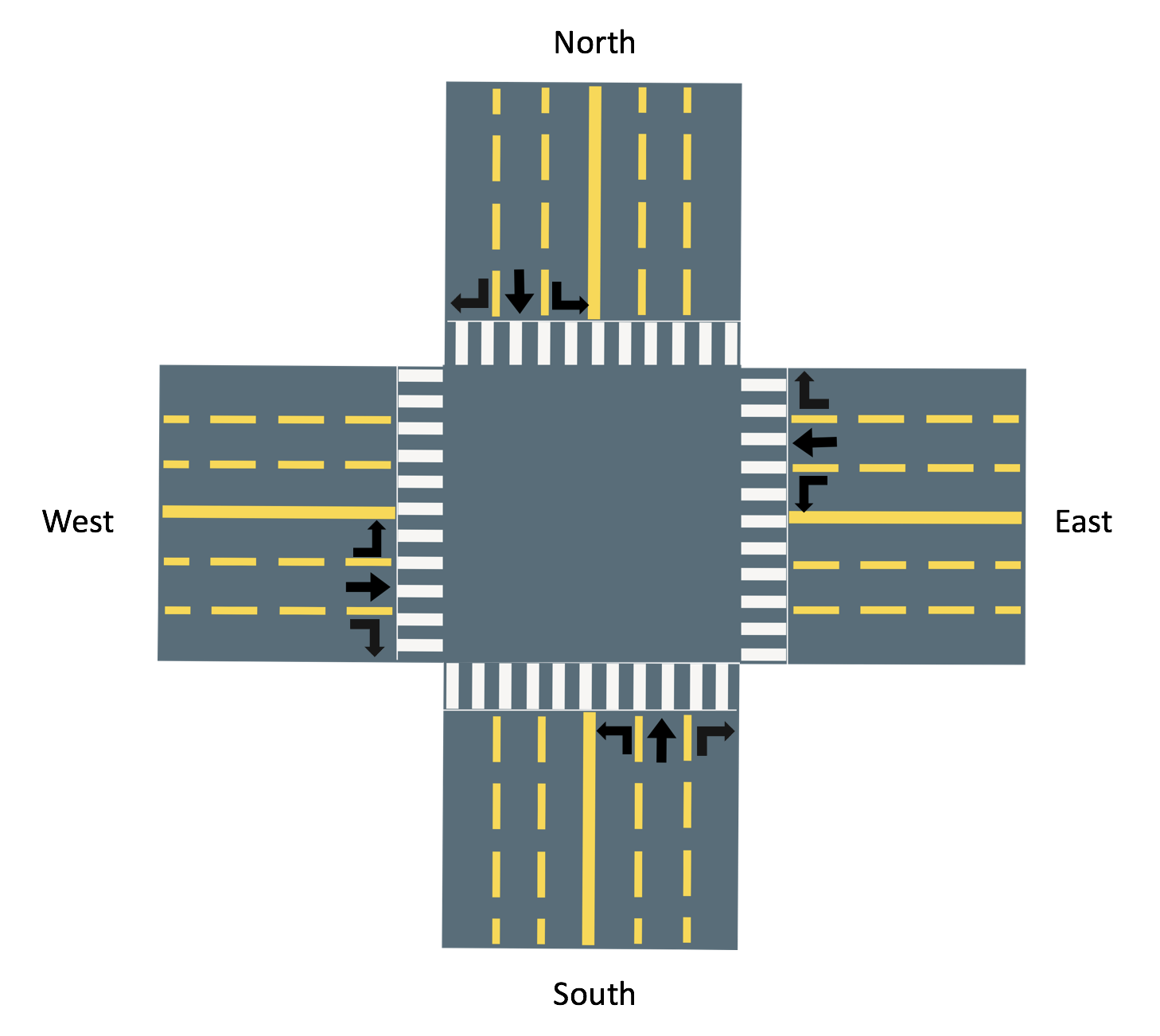}
         \caption{}
         \label{fig:a}
     \end{subfigure}
\hfill
    \begin{subfigure}[b]{0.13\textwidth}
         \centering
         \includegraphics[width=\textwidth]{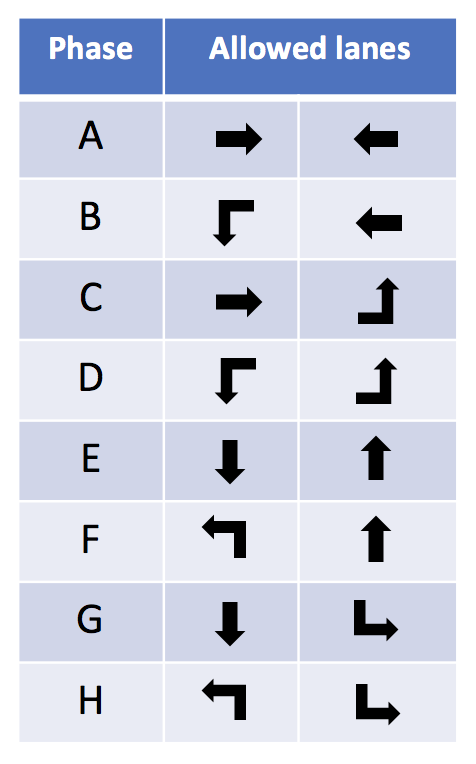}
         \caption{}
         \label{fig:b}
     \end{subfigure}
\caption{ Intersection structure and traffic signal phase adapted from [Zang et al., 2020]. (a) shows a standard intersection with four entering approaches. (b) enumerates eight typical signal phases. } 
\label{fig:intersection}
\end{figure}

Figure \ref{fig:intersection} demonstrates the a standard 4-approach intersection. Each entering approach has left-lane, through-lane and right-lane.
The control process in a intersection is modeled as a Markov decision process(MDP) $L=<S,A,R,\gamma>$ which we explain as follows:
\begin{itemize}
    \item State space $S$: Following the setting of previous work, we use current traffic flow of each lane as the states of the MDP, which includes the queue length of each lane, moving speed and so on. For a fair comparison with previous work \cite{zang2020metalight}, we use the queue length of each lane as state $S$. The number of approaches and lanes determine the dimension of state space. In real world intersections with different number of entering approaches exist(e.g. three or even five) and are regarded as heterogeneous scenarios. 
    
    \item Action space $A$: As shown in Figure \ref{fig:intersection}, there are eight signal phases in total. At each time step, the system chooses one phase among them to perform, so The dimension of action space for RL agent equals to the number of phases. Intersections may have only a subset of the eight signal phases available in action space known as phase setting \cite{zang2020metalight}.  Therefore intersections with different phase settings are regarded as heterogeneous scenarios.
    
    \item Reward $R$: To align with previous work, we use the negative sum of queue length of all lanes as the reward. We choose travel time as the evaluation metrics defined as the average travel time that vehicles spend on approaching lanes.
    
    \item Transition Probability: Different signal phase will result in different queue length after each control step. For example, a green light phase on one direction is expected to reduce the queue length along with that direction. Note that this process is stochastic because the change in queue length is also affected by the in-coming traffic which is uncertain. However, we can learn the associated probability regarding the transition in states for each action $P(s';s,a)$. This transition probability is called dynamics in RL. In this work, we use model-free RL which does not explicitly learn the dynamics, instead, it learns the value function regarding different states and actions which implicitly contains the information of dynamics.

\end{itemize}

\section{Learning}
We describe the DRL method that we apply to the single-intersection traffic signal control problem and then followed by Bayesian meta-RL.

\subsection{DQN}
Since the action space is discrete and with small dimension in traffic signal control problems, deep Q-learning(DQN) is an appropriate reinforcement learning algorithm to choose \cite{wei2019survey}. In Q-learning, the goal is to learn Q function defines as the sum of reward $r_t$ discounted by $\gamma$ at each timestep $t$:
\begin{equation}
    Q(s,a)=E[\sum_{t=0}^{\infty} \gamma^t r_t | s(0)=s,a(0)=a]
\end{equation}
and thus the optimal policy is $\pi(s) = \arg \max_a Q(s,a)$.
In DQN, the Q function is approximated by base learner $f_\theta$(which is a deep neural network in DQN) with input $s$ and a dimension $|A|$ output, each dimension corresponds to the according $Q(s,a)$. We denote the approximate Q function as $Q(s,a;\theta)$. The TD loss of neural network parameter $\theta$ is 
\begin{equation}
    \label{td-loss}
    L(\theta;D) = E_{s,a,r,s' \sim D} [(y^Q - Q(s,a;\theta) )^2 ]
\end{equation}
where $y^Q$ is the target value $r+\gamma \max_{a'} Q(s',a';\theta^-)$. $\theta^-$ is the parameter of target network that are only updated for every $i$ iterations to improve stability. Double DQN \cite{van2015deep} makes further modification and improves even more
\begin{equation} 
    y^Q \leftarrow r+\gamma  Q(s',\arg \max_a Q(s',a;\theta);\theta^-)
\end{equation}
In this paper we use Double DQN unless explicitly notified.

\subsection{Bayesian Meta-reinforcement Learning}

Following general meta-learning setting \cite{finn2017model} which assumes a task distribution $\mathcal{T}$ where all tasks are generated from. In traffic signal control, each intersection corresponds to a task. The goal of meta-learning is to design a meta-learner $M$ which takes data $D_i$ in a new scenario $i$ as input and output model parameters $\theta_i = M(D_i)$ with good performance for this scenario. Suppose $M$ is parameterized with $\alpha$, then the goal is to find $\alpha:= \arg \min_{\alpha} E_{i \sim \mathcal{T}} L_i[M(D_i;\alpha)]$, where $L_i$ is the expected loss function of task $i$. Notice that $M(;\alpha)$ does not need to be an explicit function, for example, it can be training rules with certain initialization \cite{finn2017model}. 

The meta-learning framework can be split to meta-training phase and meta-testing phase. In meta-training phase, we are given a set of training tasks $I_{train}=\{I_1,...,I_{N_{train}}\}$ sampled from task distribution $\mathcal{T}$ and the goal is to learn a good meta-learner $M(;\alpha)$ to enhance the learning efficiency of future traffic signal control tasks. In meta-testing phase, for a new traffic intersection $t$ sampled from $\mathcal{T}$, the meta-learner is input with data $D_t$ of this intersection and output the adapted model parameters $\theta_t = M(D_t;\alpha)$. The performance metrics of meta-learner is then evaluated by performance of adapted model parameters on sampled testing tasks $I_{test}=\{I_1,...,I_{N_{test}}\}$.

Within this problem setting, \cite{zou2020gradient} shows that model-agnostic Bayesian meta-learning is optimal under certain metrics. This method aims to learn a well-trained prior $P^{*}(\theta)$ of parameters in base learner $f$. Using this prior, we can get a posterior that adapts to task $i$ through Variational Inference \cite{kingma2013auto}. Model-agnostic meta-learning(MAML)(the base of MetaLight \cite{zang2020metalight}) is a special case of model-agnostic Bayesian meta-learning by using delta distributed prior $\delta_{\theta_0}(\theta)$ which weakens the robustness in adaptation and uncertainty measure as we can see in experiments. Gradient-EM Bayesian meta-learning \cite{zou2020gradient} has the advantages in traffic signal control such as computationally efficiency and allowing distributed deployment. However, it fails when directly applied to traffic signal control problem as we'll see below. This motivates us to develop the BM-DQN framework.

\section{Methodology}

In this section, we first introduce our base learner architecture BFRAP and then introduce our proposed BM-DQN framework. 

\subsection{Bayesian Base-learner Architecture}

As described above, a flexible base model $f$ is required to handle the scenario across heterogeneous intersections for different action space and state space in traffic signal control. We adopt FRAP++ \cite{zang2020metalight} and propose a Bayesian version of this structure BFRAP. 

The structures in 4-phase intersections which consists of several embedding layers and convolutional layers \cite{zang2020metalight}. The parameters of the embedding layers across different lanes are shared. The number of filters in convolutional layers are also fixed. So this structure can be applied to different phase settings and different number of lanes. 

In Bayesian learning, instead of learning a point estimator of the model parameter $\theta$, it learns a distribution $P(\theta)$ of the model parameters to handle uncertainty and to enhance robust continual learning ability. We assume Gaussian distribution of model parameter $P(\theta) = q(\theta;\lambda) \sim N(\mu_{\lambda},\sigma_{\lambda})$ in this work. 

\subsection{BM-DQN Framework}

The goal of BM-DQN is to utilize previous learned knowledge to enhance the learning process in target intersection. To do this, we meta-learn a good prior of model parameter $P^{*}(\theta)=q(\theta;\Theta)$ by alternatively perform two update steps: individual update and global update. To elaborate the details, we define inner-learner $\lambda_i$ as the distribution parameter of learned posterior over task $i$, and meta-learner $\Theta$ as the distribution parameter of meta-learned prior over the task distribution $\mathcal{T}$. During individual update, starting with some prior, the inner-learner performs Bayesian fast learning on DQN to update the posterior. During meta update, the meta-learner extracts the common knowledge over inner-learners to update the prior. 

This framework is inspired by Gradient-EM Bayesian meta-learning(GEM-BML) (see Appendix and \cite{zou2020gradient} for details). However, the original design of GEM-BML and other Bayesian meta-learning methods all focus on policy based DRL algorithms which update parameters only after each whole episode. BM-DQN improves the updating frequency by undertaking a mini-batch updating after each step in one episode which takes the advantage of fast learning in DQN. Yet, our empirical results show that the direct combination of GEM-BML and DQN does not meet our expectation. According to our analysis, it is caused by the higher update frequency requires better fast adaptation ability of the model. Therefore, we design a fast adaptation variation by adding an initial point training process during global update to enhance fast adaptation in individual update. Empirical results show that this variation largely improves the performance on traffic signal control. The framework of BM-DQN is illustrated in Algorithm \ref{algo:fvi} and we elaborate the details below.

\begin{algorithm}[htp]
\label{algo:fgem}
  \SetAlgoLined\DontPrintSemicolon
  \SetKwFunction{FGemMAML}{BM-DQN}\SetKwFunction{FGVI}{Individual-update}\SetKwFunction{BR}{BR}
  \SetKwProg{myalg}{Algorithm}{}{}
  \myalg{\FGemMAML{}}{
   randomly initialize $\Theta$, $\lambda$\;
 $t=0$\\
 \While{not done}{
  Sample batch of tasks $\mathcal{T}_t \sim P(\mathcal{T})$\;
  
  \For{each task $i \sim \mathcal{T}_t$}{
     $\lambda_i^{tr} =$\FGVI{$\Theta$,$\lambda$, $D^{tr}_i$}  \\
     $\lambda_i^{tr \oplus val} =$\FGVI{$\Theta$,$\lambda_i^{tr}$, $D^{val}_i$}
   }
   
     $\lambda  \leftarrow   
    \lambda + \gamma   \sum_{i \in \mathcal{T}_t} 
    (\lambda_i^{tr \oplus val} - \lambda)$ 
    
   $\Theta  \leftarrow \Theta - \beta \nabla_{\Theta} \big \{ KL[q(\theta;\lambda_i^{tr \oplus val}) \parallel q(\theta;\Theta)] - KL[q(\theta;\lambda_i^{tr}) \parallel q(\theta;\Theta)] \big  \}$ \\
   $t=t+1$ 
 }
 }
  \setcounter{AlgoLine}{0}
  \SetKwProg{myproc}{Subroutine}{}{}
  \myproc{\FGVI{$\Theta$, $\lambda$,$D_i$}}
  {
  Initialize $\lambda_i$ at $\lambda$.\\
  Sample $\Tilde{\epsilon}$ from $\epsilon \sim p(\epsilon)$.  \\
   $\lambda_i \leftarrow \lambda - \alpha \nabla_{\lambda} 
      [L(\theta;D_i) + KL(q(\theta_i;\lambda) \parallel p(\theta_i|\Theta))] $
 
  return $\lambda_i$
  }
  \caption{Bayesian-meta DQN(BM-DQN)}
  \label{algo:fvi}
\end{algorithm} 

\textbf{Individual update}

This step we fix the meta-learner $\Theta$ and let each inner-learner $\lambda_i$ learn from the data $D_i$ of each task $i$. As described in \cite{mnih2015human}, DQN uses a neural network with parameter $\theta$ to represent the action-state function $Q(s,a)$ in Equation (1). In traffic signal control, BFRAP follows the standard design of DQN with experience replay and target value network. In each intersection $I_i$ , the agent’s experiences $e_i (t) = (s_i (t), a_i (t), r_i (t), s_i (t + 1))$ at each timestep $t$ are stored in set $D_i$.
Then the learning process can be viewed as a Variational Inference of the posterior $q(\theta;\lambda_i)$ given the prior $q(\theta;\Theta)$. This is done by the gradient descent on the sampled ELBO loss: 
\begin{equation}
     \lambda_i \leftarrow \lambda_i - \alpha \nabla_{\lambda=\{\mu_\lambda,\sigma_\lambda\}} 
      L^{ELBO}(\mu_\lambda +  \Tilde{\epsilon} \sigma_\lambda ;D_i)
\end{equation}
where $\alpha$ is the step size, $\Tilde{\epsilon}$ is standard normal samples $\epsilon \sim N(0,1)$. The ELBO loss is the loss function L(defined in Equation (\ref{td-loss})) plus the KL-divergence between prior and posterior.
\begin{equation}
L^{ELBO}(\theta;D_i) = L(\theta;D_i) + KL[q(\theta;\lambda_i) \parallel q(\theta|\Theta)] 
\end{equation}

To speed up the learning process of this step(one gradient-step learning), we design an amortized Bayesian inference method. We start the gradient descent at a meta-learned initial point $\lambda$ instead of $\Theta$. This special initial point is commonly sensitive for the ELBO loss function surface such that one or two gradient step would be sufficient to obtain good performance on this loss.  Both $\lambda$ and $\Theta$ are updated in each Global update step based on the results in each individual update step.

\textbf{Global update} 
We then fix the posteriors and compute the update of the prior parameters $\Theta$ by equation (\ref{q}).
After the adaptation in individual-level, global-level adaptation aims to aggregate the adaptation of each intersection $I_i$ to update the initialization $\lambda$ of inner-learner and the meta-learner(prior distribution parameter) $\Theta$. To learn a good prior $\Theta$ which prevents meta-level overfitting \cite{finn2017model,zou2020gradient}, the data in individual update is split into training set and validation set. We perform individual update consecutively on both of them and obtain $\lambda_i^{tr}$ and $\lambda_i^{tr \oplus val}$. The meta-learner $\Theta$ and inner-learner initialization $\lambda$ is updated as follows:
{\small
\begin{align}
    \lambda \leftarrow   
    \lambda - & \gamma   \sum_{i \in \mathcal{T}_t} 
    (\hat{\lambda - \lambda_i^{tr \oplus val} })  \\
   \Theta  \leftarrow \Theta - & \beta \nabla_{\Theta} \big \{ KL[q(\theta;\lambda_i^{tr \oplus val}) \parallel q(\theta;\Theta)] \nonumber\\
     - & KL[q(\theta;\lambda_i^{tr}) \parallel q(\theta;\Theta)] \big  \} \label{q}
\end{align}
}%
where $\gamma,\beta$ are stepsizes. Notice that the KL divergence of two Gaussian distributions has close form solution which is differentiable. The detail of solutions is included in Appendix. Unlike traditional MAML style framework \cite{yoon2018bayesian,finn2018probabilistic,ravi2018amortized} which involves meta-update gradients over the inner-update optimization process thus leading to meta-update backProp, our method keeps the separation of meta-update and inner-update to avoid meta-update backProp while performs fast adaption. The whole algorithm for meta-training process of BM-DQN is described in Alg \ref{algo:fvi}.

\textbf{Adaptation to New Scenarios}
 In the meta-training process of BM-DQN, we learn well-generalized prior and initialization of posterior distribution on parameters $\theta$ in $f$. For a new target intersection $I_t$, we apply the learned prior parameter $\Theta$ and posterior parameter initialization $\lambda$ to it by running the individual update: 
 \begin{equation}
     \lambda_t \leftarrow {\fontfamily{qcr}\selectfont Individual-update}(\Theta,\lambda,D_t)
 \end{equation}
Then we evaluate the performance by sampling parameters $\theta$ from the task posterior distribution $q(\theta;\lambda_t)$. The meta-testing process is outlined in Alg 2.

\begin{algorithm}[t]
\small
\label{algo:fgem}
  \SetAlgoLined\DontPrintSemicolon
  \SetKwFunction{Meta}{Meta-test}\SetKwFunction{VI}{VI}
  \SetKwProg{myalg}{Algorithm}{}{}
  \myalg{\Meta{}}{
   Require: learned $\Theta, \lambda$ and $D_i^{tr}$ from new task $i$\\
   Compute posterior $\lambda_i =$\FGVI{$\Theta$,$\lambda$, $D^{tr}_i$}. \\
   Sample $\theta_i \sim P(\theta;\lambda_i)$ \\
   return $f(;\theta_i)$ for evaluation
 }
  \caption{\small Meta-testing algorithm}
 \label{algo:vi}
\end{algorithm}

\subsection{Fast Adaptation Variation}
To speed up the Gradient-based Varational Inference Subroutine, we propose a fast adaptation variation. We add an initial point training process in the meta-update phase to meta-train a good initial point $\lambda$ for the one-step gradient training of $\lambda_i$. The good initial point $\lambda$ is trained by moving it towards the two-step trained weights $\lambda_i^{tr \oplus val}$ at each meta-update step [Algorithm \ref{algo:fvi}, line 10].
Just like Reptile, this initial point $\lambda$ is optimized such that one or few number of gradient steps on the training of a new task model parameter $\lambda_i$ will result in great performance on the ELBO loss function $ELBO^{(i)}(\Theta,\lambda_i,D_i)$.  This means one or a few number of Variational Inference gradient step will quickly produce a good approximate posterior as the prior parameter $\Theta$ converges through the meta-update iterations. 

Notice that this variation maintains the advantage of GEM-BML, i.e., the separation of meta-update and inner-update to avoid meta-update backProp which is important in scaled and potentially distributed system like traffic signal control. We show the necessity of this variation by comparing BM-DQN with GEM-BML in the experiment section (ablation experiments).

\section{Related Work}

In recent decades, RL-based traffic signal control has attracted widely attention from both academia and industry. Various RL algorithms have been applied to this problem, from the early tabular Q-learning and discrete state representation methods \cite{balaji2010urban,abdulhai2003reinforcement} to current deep RL, including value based methods (e.g., deep Q-Network \cite{van2016coordinated,wei2019survey,wei2018intellilight} and policy-based methods \cite{aslani2017adaptive}.

Recently, the effort of utilizing common knowledge of intersections to enhance control has also started. A big obstacle of these direction is the lack of a universal network design for different intersection scenarios, which means that we need to train different networks for different scenarios from the start. A recently proposed a novel network design, called FRAP, introducing convolutional layers into the architecture and makes it possible to effectively share network parameters to different intersections. 

Meanwhile, meta-learning, which adapts to new task by leveraging the experience learned from similar tasks, has developed fast in recent years. In meta-reinforcement learning,
model-agnostic meta-learning (MAML) \cite{finn2017model} achieves competitive performance with good computational efficiency and becomes a leading trend. 
These development motivates MetaLight \cite{zang2020metalight}, a framework that applies FRAP and MAML into traffic signal control.  

\subsection{Bayesian Meta-learning}
Further research shows that MAML is a special case of the Hierarchical Bayesian model(HBM) \cite{grant2018recasting}, which uses Bayesian method to utilize statistical connections between related tasks. HBM have been decently studied \cite{heskes1998solving} in the past, but it is until recently it has a big comeback in deep models because of its advantages in uncertainty measure and overfitting preventing \cite{wilson2007multi}. Inspired by the training scheme of MAML, a series of MAML style Bayesian meta-learning algorithms were developed \cite{grant2018recasting,yoon2018bayesian,ravi2018amortized}. However, these methods involves optimizations over the inner-update proces during meta-update, making it hard to scale to real-world situations. Recent work GEM-BML \cite{zou2020gradient} decouples inner-update and meta-update, thus gives high flexibility to the optimization process of inner-update and making it appropriate for distributed systems.

\section{Experiment}

We first test our algorithm on a re-designed 2D navigation RL environment, then evaluate its performance in traffic signal control environment.

\subsection{Restricted 2D navigation}
To demonstrate the superiority of our algorithm in general RL environment, we design a restricted 2D navigation problem as follows. The restricted 2D navigation involves a set of tasks where a point agent must move to different goal positions in 2D, randomly chosen for each task within a unit square. The observation is the current 2D position and the actions are velocity commands in the range $[\text{-} 0.1, 0.1]$. To make this problem suitable for value-based RL algorithms like DQN, we restricted the dimension of action space to 16, which includes 8 directions, each direction with two choice of magnitude 0.03 and 0.1. The reward is the negative squared distance to the goal, and episodes terminate when the agent is within 0.02 of the goal or at the horizon of $H=100$. Detailed hyperparameter settings for this problem are in Appendix. 

In our evaluation, we compare adaptation to a new task with up to 3 episodes' training and we run the same process on 40 new tasks to calculate the average performance.  The results in Figure 3 show the adaptation performance after each episode of BM-DQN, GEM-BML and MAML. The results show that BM-DQN adapts to new task much quicker and achieves better performance than previous methods. It also ratified the necessity of our fast-adaptation variation to develop BM-DQN from GEM-BML.

\begin{figure}
    \centering
  \includegraphics[width=40mm]{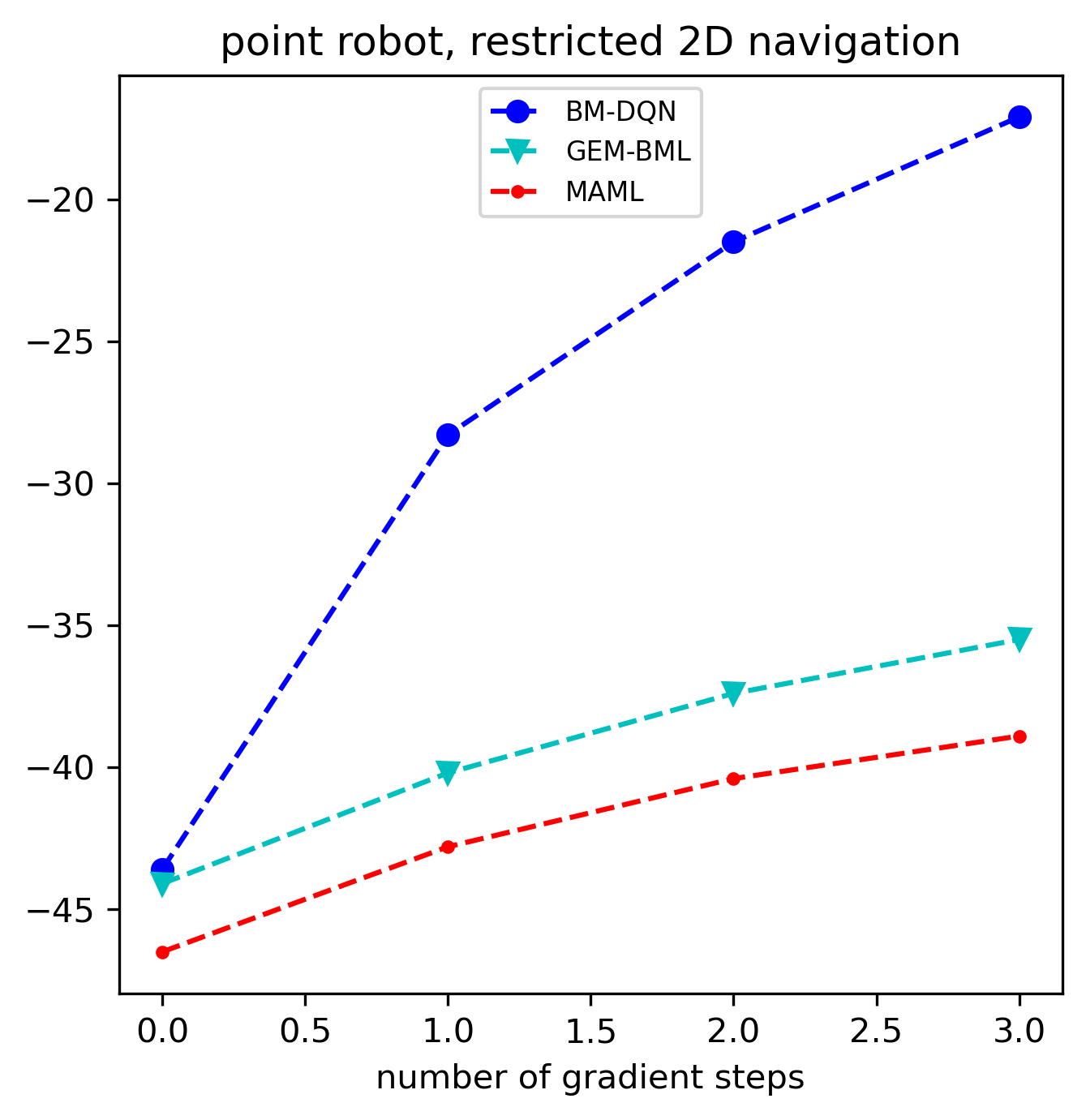}
  \caption{Meta-test results of restricted 2D navigation task} 
\end{figure}

\subsection{Traffic Signal Control}
In this experiment, we compare our BM-DQN framework with some representative benchmarks including state-of-art MetaLight \cite{zang2020metalight} (RL-based method) and Self- Organizing Traffic Light Control (SOTL) which is the representative classical transportation method \cite{cools2013self}. The results of MAML are not included since MetaLight largely outperforms MAML. We perform experiments on both an easy setting and a challenging setting. The easy setting consists of homogeneous scenarios where the tasks in meta-testing share the same phase setting and come from the same city as the tasks in meta-training. While in challenging setting, meta-testing tasks comes from different cities with different phase setting as in meta-training. We first introduce some experiment details and then represent the results of two settings. 

\subsubsection{Experiment details}
This experiments is conducted in a simulation platform called CityFlow \cite{zhang2019cityflow}, one of the latest simulation environments for traffic signal control. To use the simulator, we first feed real traffic datasets into the platform then we can use the simulator as a RL environment which executes the actions and returns the state and reward in each time step. 

We use the datasets provided in \cite{zang2020metalight} which includes four real-world datasets from two cities in China: Ji- nan (JN) and Hangzhou (HZ), and two cities in the United States: Atlanta (AT), and Los Angeles (LA). The raw traffic data from two Chinese cities contains the information about the vehicles coming through the intersections, which are captured by the nearby surveillance cameras. The other raw data from American cities is composed of the full vehicle trajectories which are collected by several video cameras along the streets3. Just like most intersections, the entering lanes only consist of left-lane and through-lane. The episode length is one hour. In order to build the simple and challenging settings as elaborated below, different phase settings are created. The details of phase settings are summarized in Appendix. 

We compare our BM-DQN model with MetaLight \cite{zang2020metalight} and GEM-BML \cite{zou2020gradient}. For fair comparison, all models use the same Double DQN update algorithm and the same nerual network architecture FRAP as illustrated in Figure 2.

We choose travel time as the evaluation metric which is defined as the average travel time that vehicles spend on approaching lanes. More details such as hyperparameter are in Appendix.

\subsubsection{Simple Settng: Homogeneous Scenarios}

In this setting, we first perform meta-training on 25 tasks from datasets of Hangzhou.
During meta-testing, we choose six homogeneous tasks also from datasets of Hangzhou whose phase settings exist in the meta-training set.
The results are described in Table 1 and Figure 4. Each phase setting stands for one task. Notice that in table 1 the average travel time among the whole adaptation process is reported, while in \cite{zang2020metalight} the minimum travel time is reported. We choose these metrics because we focus on the robustness in the whole adaptation process. In Figure 4 we plot the adaptation curve during meta-testing. The model is trained in DQN scheme and we perform evaluation after each update step.  The results show that our model performs better than previous work in most scenarios with better stability, but the improvement is not significant. It is within our expectation because in homogeneous setting the common knowledge of training tasks in this setting is sufficient for quick adaptation on new tasks by using point estimated meta-knowledge. 

We can see that GEM-BML cannot keep a stable learning trend or adapt slowly. It is because of the high update frequency in DQN training scheme that the adaptation speed of the inner-learners can not follow up without our fast-adpatation variation. In contrast, BM-DQN maintains a more stable and faster adaptation.

\begin{figure} 
    \centering
  \includegraphics[width=90mm]{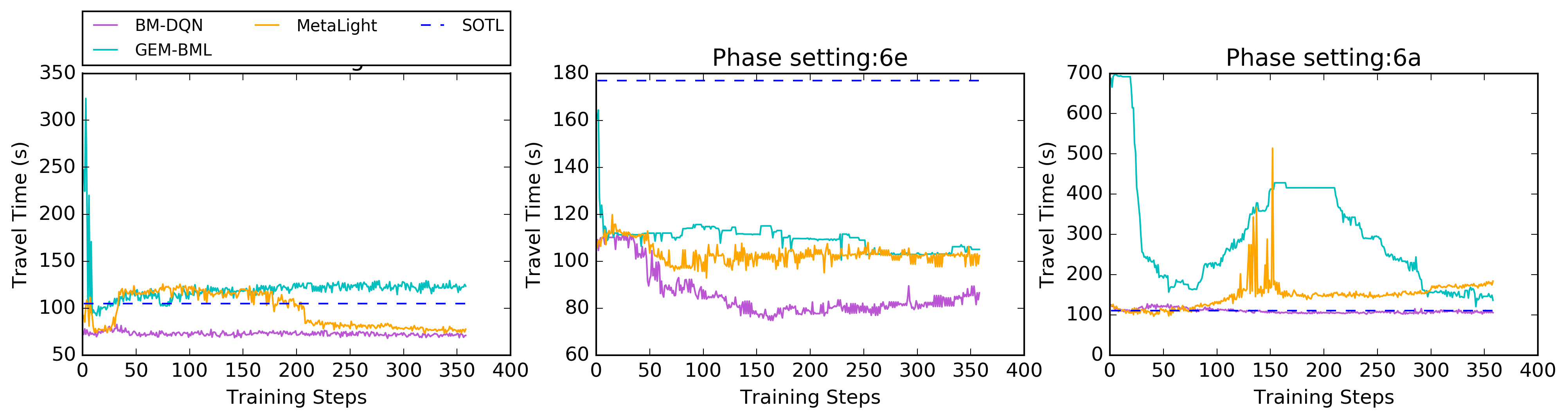}
  \caption{Meta-testing results of simple settings. Each point is regarding a whole episode evaluation after each DQN training step.} 
\end{figure}

\begin{table}[t]

\centering
\begin{tabular}{ |c|c|c|c| } 
 \hline
 Phase Setting & 8 &  6e & 6a\\ 
 \hline
 GEM-BML &  119.98  & 109.57  & 294.82  \\ 
 \hline
 MetaLight &  97.48  & 103.13  & 148.09  \\ 
 \hline
 BM-DQN & 72.72   & 85.85  & 109.36 \\ 
 \hline
\end{tabular}
\caption{Results of different methods on simple setting. Average travel time among the whole adaptation process is reported. }
\label{table:image}
\end{table}

\subsubsection{Challenging Settng: Heterogeneous Scenarios}

\begin{table}[t]
\tiny
\centering
\begin{tabular}{ |c|c|c|c|c|c|c| } 
 \hline
 Phase Setting & LA-2 &  Atlanta-2 & LA-1 & Atlanta-1 & Jinan-1 & Jinan-2\\ 
 \hline
 GEM-BML &  238.89  & 777.89  & 666.94  & 457.01 & 153.61 & 535.75 \\ 
 \hline
 MetaLight &  257.67  & 898.91 & 989.77 & 494.36 & 173.26 &   587.27\\ 
 \hline
 BM-DQN & 135.14   & 675.66 & 885.27  & 192.01 & 119.50 &   549.69\\ 
 \hline
\end{tabular}
\caption{Results of different methods on challenging setting. Average travel time among the whole adaptation process is reported. }
\label{table:image}
\end{table}

\begin{figure} 
    \centering
  \includegraphics[width=80mm]{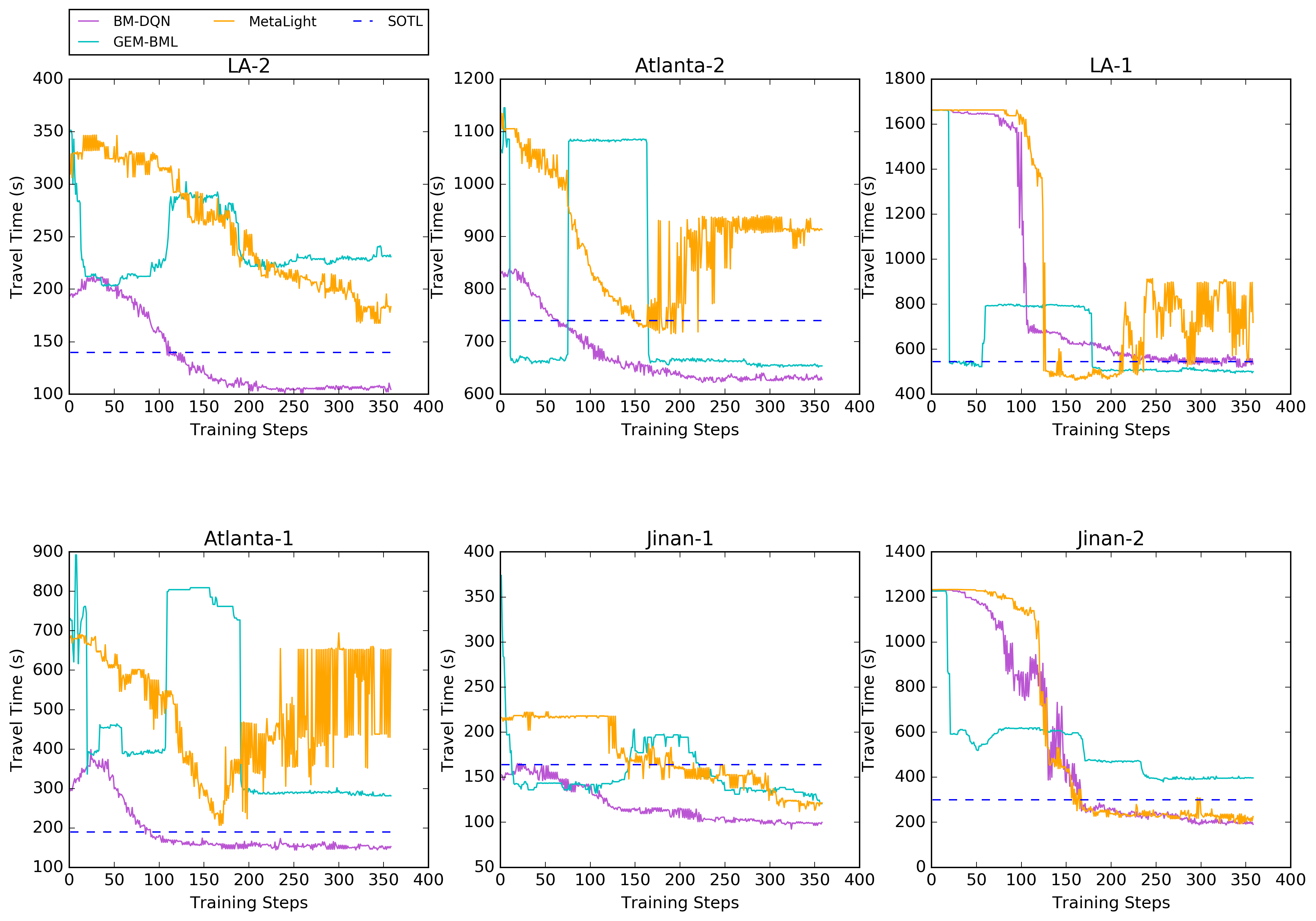}
  \caption{Meta-testing results of challenging settings} 
\end{figure}

In this setting, we try to test the knowledge transfer ability not just between different phase settings but also between different cities. As described in \cite{zang2020metalight}, the source data may differ greatly between cities, which increasing the difficulties to adapt control policy. 

For meta-training, we use the same training tasks from datasets of Hangzhou as in the simple setting.
While in meta-testing we perform hetero-geneous adaptation on tasks from datasets of Jinan, Atlanta, and Los Angeles. The results are presented in Table 2 and Figure 5. Compared with simple setting, BM-DQN significantly outperforms all baselines and adapts much faster and more stable. It is because  under this setting the new tasks in meta-testing have a relatively different distribution from the training tasks in meta-training, which is more likely to cause meta-level overfitting \cite{yoon2018bayesian}. 
Since BM-DQN learns an effective probabilistic prior knowledge, it could enable more robust continual learning in new scenarios and mitigate the impact of meta-level overfitting.

\clearpage
\bibliographystyle{named}
\bibliography{ms}

\end{document}


\onecolumn
\icmltitle{Appendix}



\icmlsetsymbol{equal}{*}

\icmlaffiliation{to}{Department of Computation, University of Torontoland, Torontoland, Canada}
\icmlaffiliation{goo}{Googol ShallowMind, New London, Michigan, USA}
\icmlaffiliation{ed}{School of Computation, University of Edenborrow, Edenborrow, United Kingdom}

\icmlcorrespondingauthor{Cieua Vvvvv}{c.vvvvv@googol.com}
\icmlcorrespondingauthor{Eee Pppp}{ep@eden.co.uk}

\icmlkeywords{Machine Learning, ICML}

\vskip 0.3in




\appendix




\section{Preliminary of Bayesian Meta-learning}

Consider a set of related tasks $\tau \in \mathcal{T}$, each task is specified by a dataset $D_\tau$ generated by its underlying model $P(D_\tau|\theta_\tau)$. We assume the tasks are statistically related such that their model parameters are sampled from a task distribution $\theta_\tau \sim p(\theta_\tau|\Theta)=P(\mathcal{T})$, where $\Theta$ is the prior distribution parameters. In Bayesian Hierarchical Models such as random-effect model, Maximum Likelihood Method is performed to estimate 
\begin{align} \label{eq0}
    \hat{\Theta}=\arg\max_{\Theta} \prod_{\tau \in \mathcal{T}} P(D_\tau|\Theta) 
\end{align}
where $P(D_\tau|\Theta)=\int p(D_\tau|\theta_\tau)p(\theta_\tau|\Theta)d\theta_\tau$ .

 Instead of optimizing the likelihood of the task prior as (\ref{eq0}), GEM-BML \cite{zou2020gradient} optimizes the likelihood of the task posterior $L(\Theta,D^{tr}_\mathcal{T};D^{val}_\mathcal{T}) = \log P(D^{val}_\mathcal{T}|\Theta,D^{tr}_\mathcal{T})$ in order to prevent overfitting:
\begin{equation} \label{eq1}
    \hat{\Theta}=\arg\max_{\Theta} \prod_{\tau \in \mathcal{T}} P(D_\tau^{val}|\Theta,D^{tr}_\tau)
\end{equation}
where \begin{equation} \label{eq2} P(D_\tau^{val}|\Theta,D^{tr}_\tau)=\int p(D^{val}_\tau|\theta_\tau)p(\theta_\tau|\Theta,D^{tr}_\tau)d\theta_\tau.  
\end{equation}

GEM-BML is based on the two following lemmas.
The first lemma is inspired by the observation in \cite{salakhutdinov2003optimization}  that by knowing the posterior one can compute the exact gradient of the task-prior likelihood $L(\Theta;D_\mathcal{T}) = \log P(D_\mathcal{T}|\Theta)$ through gradient of the \textit{complete} log likelihood. 

\begin{lemma}
\label{cor1}
\begin{align*}
&\nabla_{\Theta} L(\Theta;D_\mathcal{T})
\\ &=\sum_{\tau \in \mathcal{T}}  E_{\theta_\tau|\Theta, D_\tau } \nabla_{\Theta} \log [ P(D_\tau|\theta_\tau)* P(\theta_\tau|\Theta) ]
\\ &=\sum_{\tau \in \mathcal{T}}  E_{\theta_\tau|\Theta, D_\tau } \nabla_{\Theta} \log P(\theta_\tau|\Theta)
\end{align*}
\end{lemma}
where $P(\theta_\tau|\Theta)$ is the prior and $E_{\theta_\tau|\Theta, D_\tau }$ means the expectation of $\theta_\tau$ under the posterior $P(\theta_\tau|\Theta, D_\tau )$. 

The second lemma shows that the task-posterior likelihood can be written as the difference of the two task-prior likelihoods of $D^{tr}_\mathcal{T}$ and $D^{tr}_\mathcal{T} \bigcup D^{val}_\mathcal{T}$:

\begin{lemma}
\label{cor2}

$L(\Theta,D^{tr}_\mathcal{T};D^{val}_\mathcal{T}) = L(\Theta;D^{tr}_\mathcal{T} \bigcup D^{val}_\mathcal{T}) - L(\Theta;D^{tr}_\mathcal{T})$
\end{lemma}
Combining Lemma \ref{cor1} and \ref{cor2} we obtain
\begin{align} 
&\nabla_{\Theta} L(\Theta,D^{tr}_\mathcal{T};D^{val}_\mathcal{T})  \nonumber
\\ = & \sum_{\tau \in \mathcal{T}}  E_{\theta_\tau|\Theta, D^{tr}_\tau \bigcup D^{val}_\tau } \nabla_{\Theta} \log P(\theta_\tau|\Theta) \nonumber \\ &  
  -  E_{\theta_\tau|\Theta, D^{tr}_\tau } \nabla_{\Theta} \log P(\theta_\tau|\Theta) \label{eq}
\end{align}
This equation leads to an iterative algorithm. In individual-update, we first fix the prior parameters $\Theta$ and compute the corresponding posteriors $P(\{\theta_\tau\}|\Theta,D^{tr}_\mathcal{T})$ and $P(\{\theta_\tau\}|\Theta,D^{tr}_\mathcal{T}\bigcup D^{val}_\mathcal{T})$. In global-update, we fix the posteriors and compute the update of the prior parameters $\Theta$ by equation (\ref{eq}). By iteratively performing the individual-update and the global-update we are performing gradient ascent of posterior likelihood $L(\Theta,D^{tr}_\mathcal{T};D^{val}_\mathcal{T})$.

\section{Solution details of global update}

Under Gaussian approximation, we assume the prior and approximate posteriors to be $q(\theta_\tau|\Theta) \sim N(\mu_{\Theta},\Lambda_{\Theta}^{-1})$ and 
$q(\theta_\tau;\lambda^{tr}_\tau) \sim N(\mu_{\theta_\tau}^{tr},\Lambda_{\theta_\tau}^{tr})$, $q(\theta_\tau;\lambda^{tr\oplus val}_\tau) \sim N(\mu_{\theta_\tau}^{tr\oplus val},\Lambda_{\theta_\tau}^{tr\oplus val})$. Then $\nabla_{\Theta} \big \{ KL[q(\theta;\lambda_i^{tr \oplus val}) \parallel q(\theta;\Theta)] - KL[q(\theta;\lambda_i^{tr}) \parallel q(\theta;\Theta)] \big  \}$ in equation (4) Section 3.1 has close form solution given as follows. 

$$\frac{\partial \big \{ KL[q(\theta;\lambda_i^{tr \oplus val}) \parallel q(\theta;\Theta)] - KL[q(\theta;\lambda_i^{tr}) \parallel q(\theta;\Theta)] \big  \} }{\partial \mu_{\Theta}} 
= \sum_{\tau \in \mathcal{T}} (\mu_{\theta_\tau}^{tr\oplus val} - \mu_{\theta_\tau}^{tr})^T \Lambda_{\Theta}^{-1}$$
\begin{equation}
\begin{split}
&\frac{\partial \big \{ KL[q(\theta;\lambda_i^{tr \oplus val}) \parallel q(\theta;\Theta)] - KL[q(\theta;\lambda_i^{tr}) \parallel q(\theta;\Theta)] \big  \} }{\partial \Lambda_{\Theta}^{-1}} 
= \sum_{\tau \in \mathcal{T}} - \frac{1} {2} ( \Lambda_{\theta_\tau}^{tr\oplus val} -  \Lambda_{\theta_\tau}^{tr} ) \\
&-  \frac{1} {2} (\mu_{\theta_\tau}^{tr\oplus val}-\mu_{\theta_\tau}^{tr}) 
(\mu_{\theta_\tau}^{tr\oplus val} + \mu_{\theta_\tau}^{tr} - 2\mu_{\Theta})^T
\end{split}
\label{gs}
\end{equation}

\section{Hyperparameters for restricted 2D navigation }

\begin{table}[H]
\centering
\begin{tabular}{ |c|c| } 
\hline
Global-update Learning Rate & 0.001  \\ 
 \hline
Individual-update Learning Rate & 0.1  \\ 
  \hline
  Episode Length & 100 \\ 
 \hline
 Meta-batch Size  & 20    \\
 \hline
\end{tabular}
\caption{Hyperparameters for restricted 2D navigation. }
\label{table:rl}
\end{table}

\section{Details of phase settings in Traffic Signal Control Experiment}

\begin{table}[H]
\centering
\begin{tabular}{ |c|c|c|c|c|c|c|c|c| } 
 \hline
 & A & B & C &D &E&F &G & H \\
 \hline
8 & \checkmark & \checkmark & \checkmark &\checkmark & \checkmark & \checkmark & \checkmark & \checkmark  \\ 
 \hline
6a & \checkmark & \checkmark & \checkmark &\checkmark & \checkmark &  & & \checkmark  \\ 
  \hline
6e & \checkmark & \checkmark & \checkmark &\checkmark & \checkmark &  &\checkmark  &  \\  
 \hline
LA-1 & \checkmark &  & \checkmark &\checkmark &  & \checkmark & \checkmark & \checkmark  \\ 
 \hline
LA-2  & \checkmark &  &  &\checkmark & &\checkmark &\checkmark &     \\
 \hline
 Atlanta-1  &  & \checkmark & \checkmark & &\checkmark & &  & \checkmark    \\
 \hline
 Atlanta-2  & \checkmark &  &  &\checkmark & &\checkmark &\checkmark &     \\
 \hline
 Jinan-1  & \checkmark &  &  &\checkmark & &\checkmark &\checkmark &     \\
 \hline
 Jinan-2  & \checkmark &  & \checkmark &\checkmark &  & \checkmark & \checkmark & \checkmark  \\ 
 \hline
\end{tabular}
\caption{Details of phase settings }
\end{table}

\section{Hyperparameter Settings in Traffic Signal Control Experiment}

Meta-learner updates itself every 10 time steps while inner-learners update themselves every time step.

\begin{table}[H]
\centering
\begin{tabular}{ |c|c| } 
 \hline
Global-update Learning Rate & 0.001  \\ 
 \hline
Individual-update Learning Rate & 0.001  \\ 
  \hline
Episode Length & 360 \\ 
 \hline
Individual Gradient steps at meta-test & 3 \\ 
 \hline
 Meta-batch Size  & 30    \\
 \hline
\end{tabular}
\caption{Hyperparameters in Traffic Signal Control Experiment. }
\label{table:rl}
\end{table}

{\small
\bibliographystyle{icml2019}
\bibliography{ms}
}